\documentclass[journal]{IEEEtran}
\usepackage{cite}

\ifCLASSINFOpdf
  \usepackage[pdftex]{graphicx}
  \graphicspath{{./figure/}{./}}
  \DeclareGraphicsExtensions{.pdf}
\else
  \usepackage[dvips]{graphicx}
  \graphicspath{{./figure/}{./}}
  \DeclareGraphicsExtensions{.pdf}
\fi
\usepackage{amsmath}
\usepackage{url}

\newcommand*{\adj}{^{\mathsf{H}}}

\begin{document}
\title{Highly Scalable Image Reconstruction using Deep Neural Networks with Bandpass Filtering}
\author{Joseph~Y.~Cheng,
  Feiyu~Chen,
  Marcus~T.~Alley,
  John~M.~Pauly,~\IEEEmembership{Member,~IEEE,}
  and~Shreyas~S.~Vasanawala
\thanks{J.Y. Cheng, M.T. Alley, and S.S. Vasanawala are with the Department of Radiology, Stanford University, Stanford, CA, USA email: jycheng@stanford.edu}
\thanks{F. Chen and J.M. Pauly are with the Department of Electrical Engineering, Stanford University, Stanford, CA, USA.}}

%



\maketitle

\begin{abstract}
  To increase the flexibility and scalability of deep neural networks for image reconstruction, a framework is proposed based on bandpass filtering. For many applications, sensing measurements are performed indirectly. For example, in magnetic resonance imaging, data are sampled in the frequency domain. The introduction of bandpass filtering enables leveraging known imaging physics while ensuring that the final reconstruction is consistent with actual measurements to maintain reconstruction accuracy. We demonstrate this flexible architecture for reconstructing subsampled datasets of MRI scans. The resulting high subsampling rates increase the speed of MRI acquisitions and enable the visualization rapid hemodynamics.
\end{abstract}


%
\IEEEpeerreviewmaketitle

\section{Introduction}
%
%
%
%
\IEEEPARstart{C}{onvolutional} neural network (CNN) is a powerfully flexible tool for computer vision and image processing applications. Conventionally, CNNs are trained and applied in the image domain. With the fundamental elements of the network as simple convolutions, CNNs are simple to train and fast to apply. The intensive processing can be easily reduced by focusing on localized image patches. CNNs can be trained on smaller images patches while still allowing for the networks to be applied to the entire image without any loss of accuracy.

For applications where image data are indirectly collected, this scalability and flexibility of CNNs are lost. As a specific example, we focus our proposed approach on magnetic resonance imaging (MRI) where the data acquisition is performed in the frequency domain, or k-space domain. For MRI, data at only a single k-space location can be measured at any given time; this process results in long acquisition times. Scan times can be reduced by simply subsampling the acquisition. Being able to reconstruct MR images from vastly subsampled acquisitions has significant clinical impact by increasing the speed of MRI scans and enabling visualization of rapid hemodynamics \cite{Vasanawala2010a}. Using advanced reconstruction algorithms, images can be reconstructed with negligible loss in image quality despite high subsampling factors ($>8$ over Nyquist). To achieve this performance, these algorithms exploit the data acquisition model with the localized sensitivity profiles of high-density receiver coil arrays for ``parallel imaging'' \cite{Griswold2002,Lustig2010,Pruessmann1999,Sodickson1997,Uecker2013}.
Also, image sparsity can be leveraged to constrain the reconstruction problem for compressed sensing \cite{Candes2006,Donoho2006,Lustig2007a}. With the use of nonlinear sparsity priors, these reconstructions are performed using iterative solvers \cite{Beck2009,Boyd2010,Goldstein2009a}. Though effective, these algorithms are time consuming and are sensitive to tuning parameters which limit their clinical utility.

We propose to use CNNs for image reconstruction from subsampled acquisitions in the spatial-frequency domain, and transform this approach to become more tractable through the use of bandpass filtering. The goal of this work is to enable an additional degree of freedom in optimizing the computation speed of reconstruction algorithms without compromising reconstruction accuracy. This hybrid domain offers the ability to exploit localized properties in both the spatial and frequency domains. More importantly, if the sensing measurement is in the frequency domain, this architecture enables simple parallelization and allows for scalability for applying deep learning algorithms to higher and multi-dimensional space.

\begin{figure*}[t]
  \centering
  \includegraphics[width=0.8\linewidth]{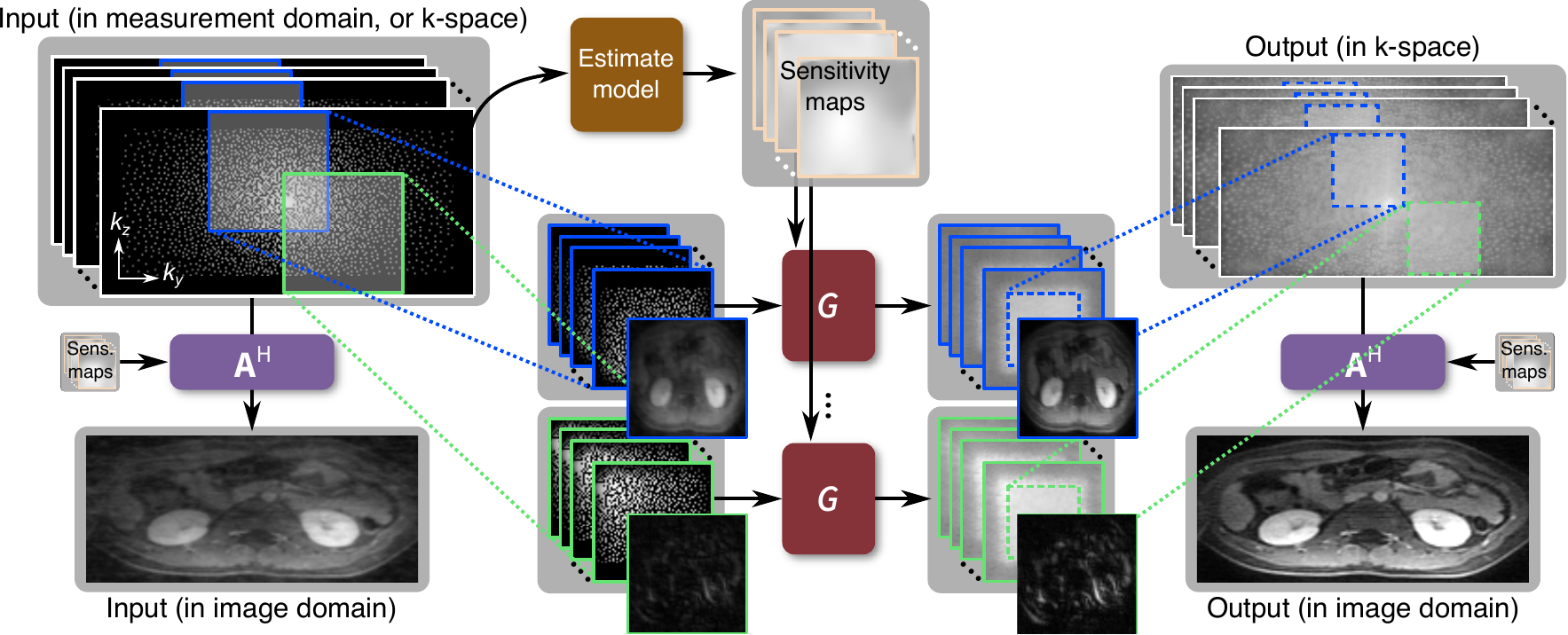}
  \caption{Method overview. We start with subsampled multi-channel measurement data in the k-space domain. The imaging model is first estimated by extracting the sensitivity maps of the imaging sensors specific for the input data. This model can be directly applied with the model adjoint $\mathbf{A}\adj$ operation to yield a simple image reconstruction (lower left) with image artifacts from data subsampling. In the proposed method, a patch of the input k-space data is inserted into a deep neural network $G$, which also uses the imaging model in the form of sensitivity maps. The output of $G$ is a fully sampled patch for that k-space region. This patch is then inserted into the final k-space output. Two example patches are shown in blue and green with corresponding images overlaid. By applying this network for all k-space patches, the full k-space data is reconstructed (upper right). The final artifact-free image is shown in the lower right.}
  \label{fig:method:overview}
\end{figure*}

\section{Related Work}
Deep neural networks have been designed as a compelling alternative to traditional iterative solvers for reconstruction problems \cite{Adler2017,Diamond2017,Hammernik2017,Jin2016,Lee2017a,Quan2017,Wurfl2016,Yang2016a}. Tuning parameters for conventional solvers, such as regularization parameters and step sizes, are learned during training of these networks which increases the robustness of the final image reconstruction algorithm. Adjustable parameters, such as learning rates, only need to be determined and set during the training phase. Also, these networks have a fixed structure and depth, and the networks are trained to converge after this fixed depth. This set depth limits the computational complexity of the reconstruction with little to no loss in image quality. Further, computational hardware devices are optimized to rapidly perform the fundamental operations in a neural network.

Three main obstacles limit the use of CNNs for general image reconstruction. First, previously proposed networks do not explicitly enforce that the output will not deviate from the measured data \cite{Lee2017a,Quan2017}. Without a data consistency step, deep networks may create or remove critical anatomical and pathological structures, leading to erroneous diagnosis. Second, if the measurement domain is not the same domain as where the CNN is applied (such as in the image domain) and a data consistency step is used, the training and inference can no longer be patch based. If only a small image patch is used, known information in the measurement domain (k-space domain for MRI) is lost. As a result, CNNs must be trained and applied on fixed image dimensions and resolutions \cite{Diamond2017,Hammernik2017,Jin2016,Wurfl2016,Yang2016a}. This limitation increases memory requirements and decreases speed of training and inference. Lastly, parallelization of the training and inference of the CNN is not straightforward: specific steps within the CNN (such as transforming from k-space domain to image domain) require gathering all data before proceeding. To address these limitations, we introduce a generalized neural network architecture.

Here, we develop an approach for image reconstruction with deep neural networks applied to patches of data in the frequency domain. In other words, a bandpass filter is used to select and isolate the reconstruction to small localized patches in the frequency space, or k-space. Previously, Kang et~al demonstrated effective de-noising with CNNs in the Wavelet domain for low-dose CT imaging \cite{Kang2016a}. Here, we extend that concept to be applicable to any frequency band, and we explicitly leverage the physical imaging model. With contiguous patches of k-space, we maintain the ability to apply the data acquisition model which enables a network architecture to enforce consistency with the measured data. Also, by selecting small patches of k-space domain, the input dimensions of the networks are reduced which decreases memory footprint and increases computational speed. Thus, the possible resolutions are not limited by the computation hardware or the acceptable computation duration for high-speed applications. Lastly, each k-space patch can be reconstructed independently which enables simple parallelization of the algorithm and further increases computational speed. With the described method, deep neural networks can be applied and trained on images with high dimensions ($>256$) and/or multiple dimensions (3+ dimensions) for a wide range of applications.


\section{Method}
\subsection{Reconstruction Overview}
\begin{figure}
  \centering
  \includegraphics[width=0.8\linewidth]{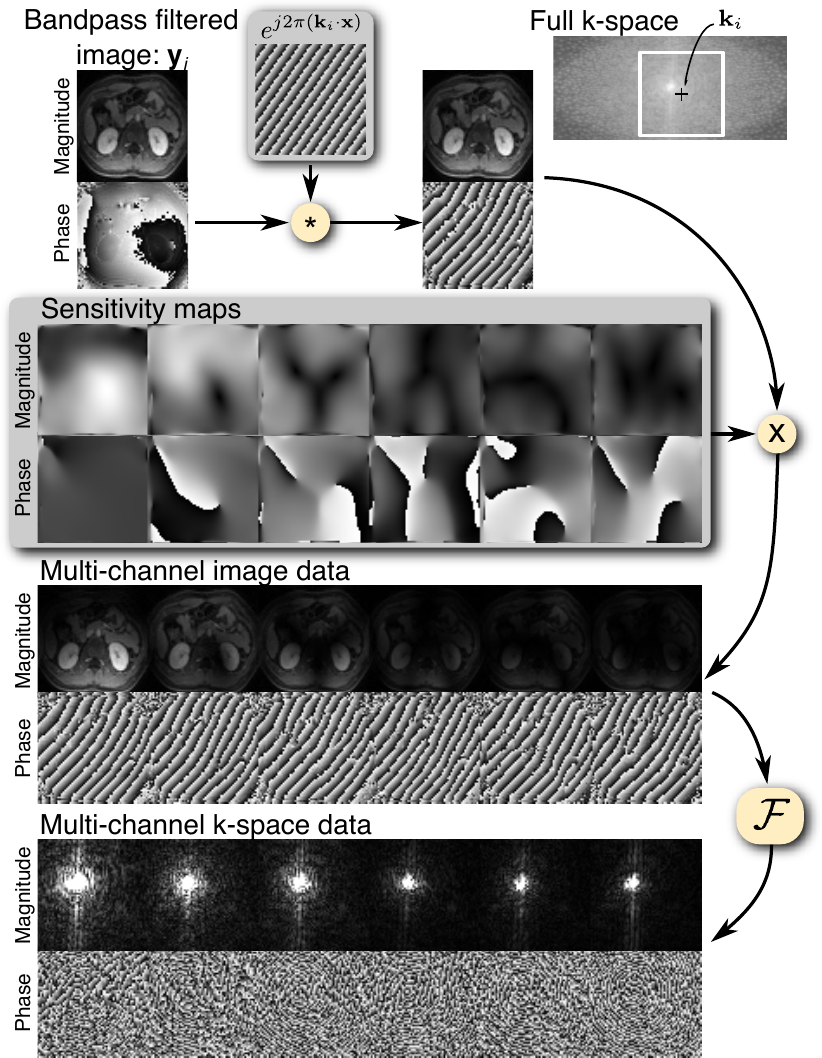}
  \caption{MRI model applied to a bandpass-filtered image. Image $\mathbf{y}_i$ corresponds to a bandpass-filtered image where a windowing function centered at $\mathbf{k}_i$ was applied in frequency space (or k-space). First, a phase modulation $e^{j\pi(\mathbf{k}_i\cdot\mathbf{x})}$ is applied to the image through a point-wise multiplication ($\ast$). The image is then multiplied by the sensitivity maps to yield multi-channel data. A Fourier transform operator $\mathcal{F}$ transforms the data to k-space.}
  \label{fig:method:model}
\end{figure}

Training and inference are performed on localized patches of k-space as illustrated in Fig.~\ref{fig:method:overview}. For the $i$-th localized k-space patch, data acquisition can be modeled as:
\begin{equation}
	\mathbf{u}_i = \mathbf{M}_i \mathbf{A} \left( e^{j2\pi(\mathbf{k}_i\cdot\mathbf{x})} \ast \mathbf{y}_i \right). \label{eq:model}
\end{equation}
The imaging model is represented by $\mathbf{A}$ which transforms the desired image $\mathbf{y}_i$ to the measurement domain. For MRI, this imaging model consists of applying the sensitivity profile maps $\mathbf{S}$ and applying the Fourier transform $\mathcal{F}$ to transform the image to the k-space domain. Sensitivity maps $\mathbf{S}$ are independent of the k-space patch location and can be estimated using conventional algorithms, such as ESPIRiT \cite{Uecker2013}. Since $\mathbf{S}$ is set to have the same image dimensions as the k-space patch, $\mathbf{S}$ is faster to estimate and have a smaller memory requirement in this bandpass formulation. This imaging model is illustrated in Fig.~\ref{fig:method:model}.

Matrix $\mathbf{M}_i$ is then applied to mask out the missing points in the selected patch $\mathbf{u}_i$. When selecting the k-space patch of $\mathbf{u}_i$ with its center pixel at k-space location $\mathbf{k}_i$, a phase is induced in the image domain. To remove the impact of this phase when solving the inverse problem, the phase is modeled separately as $e^{j2\pi(\mathbf{k}_i\cdot\mathbf{x})}$ where $\mathbf{x}$ is the corresponding spatial location of each pixel in $\mathbf{y}_i$, and $j=\sqrt{-1}$. This phase is applied through an element-wise multiplication, denoted as $\ast$.
With any standard algorithm for inverse problems \cite{Beck2009,Boyd2010,Goldstein2009a}, $\mathbf{y}_i$ from (\ref{eq:model}) can be estimated as $\mathbf{\hat{y}}_i$ through a least-squares formulation with a regularization function $R\left(\mathbf{y}_i\right)$ and parameter $\lambda$:
\begin{multline}
	\mathbf{\hat{y}}_i = \arg\min_{\mathbf{y}_i} \left\|\mathbf{W} \left[ \mathbf{M}_i \mathbf{A} \left( e^{j2\pi(\mathbf{k}_i\cdot\mathbf{x})} \ast \mathbf{y}_i \right) -  \mathbf{u}_i\right]\right\|^2_2 \\ + \lambda R\left(\mathbf{y}_i\right).\label{eq:leastsquare}
\end{multline}
In (\ref{eq:leastsquare}), we introduce a windowing function $\mathbf{W}$ to avoid Gibbs ringing artifacts when the patch dimension is too small ($<128$). The model $\mathbf{A}$ includes sensitivity maps $\mathbf{S}$ that can be considered as a element-wise multiplication in the image domain or a convolution in the k-space domain. This window function also accounts for the wrapping effect of the k-space convolution when applying $\mathbf{S}$ in the image domain. In our experiments, $\textbf{W}$ was designed as a rectangle convolved with a gaussian window for a stopband of 10 pixels. Input k-space data were first zero-padded with 10 pixels before the patch-based reconstruction to account for the stopband.



Incorporating a strong prior in the form of regularization has been demonstrated to enable high image quality despite high subsampling factors. In compressed sensing, the sparsity of the image in a transform domain, such as spatial Wavelets or finite differences, can be exploited to enable subsampling factors over 8 times Nyquist rates \cite{Lustig2007a,Otazo2010a}. Even though our problem formulation is similar to applying Wavelet transforms, directly enforcing sparsity in that domain may not be the optimal solution, and regularization parameters for each k-space location must be tuned. Thus, instead of solving (\ref{eq:leastsquare}) using a standard algorithm, we will be leveraging deep neural networks. The idea is that these networks can be trained to rapidly solve the many small inverse problems in a feed-forward fashion. Based on the input k-space patch, the network should be sufficiently flexible to adapt to solve the corresponding inverse problem. This deep learning approach can be considered as learning a better de-noising operation for each specific bandpass-filtered image for a stronger image prior.

After different frequency bands are reconstructed, the k-space patches are gathered to form the final image. The setup allows for flexibility in choosing patch dimensions and amount of overlap between each patch. These parameters were explored in our experiments. In the areas of overlap, outputs were averaged for the final solution.


\subsection{Network Architecture}
\begin{figure}[!t]
  \begin{center}
    \includegraphics[width=\linewidth]{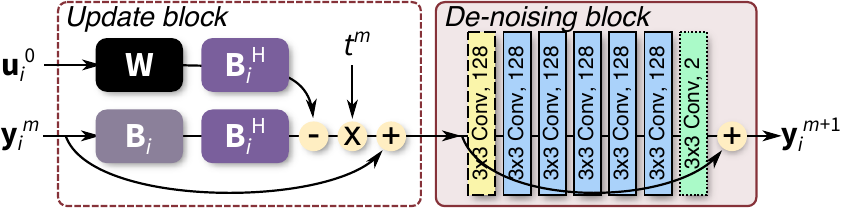}
  \end{center}
  \caption{Single ``iteration'' of the proposed network with two blocks: update block and de-noising block. The $m$-th step is illustrated to update current estimate $\textbf{y}_i^m$ to $\textbf{y}_i^{m+1}$ for the $i$-th k-space patch. In the de-noising block, an initial convolutional layer (dashed yellow) transforms the 2-channel data (real and imaginary) to 128 feature maps. Each convolutional layer is followed by batch normalization and ReLU activation except for the last layer (dotted green).}
  \label{fig:method:network}
\end{figure}

We propose to solve the inverse problem of (\ref{eq:leastsquare}) with a deep neural network, denoted as $G(.)$ in Fig.~\ref{fig:method:overview}. Any network architecture can be used for this purpose. To demonstrate the ability to incorporate known imaging physics, the architecture used is based on an unrolled optimization with deep priors \cite{Diamond2017}.
More specifically, we structured the network architecture based on the iterative soft-shrinkage algorithm (ISTA) \cite{Daubechies2004,Elad2007,Figueiredo2003,Starck2005} as illustrated in Fig.~\ref{fig:method:network}. In this framework, two different blocks are repeated: 1) update block and 2) de-noising block (or soft-shrinkage block).

The \textit{update block} enforces consistency with the measured data samples. This block is critical to ensure that the final reconstructed image agrees with the measured data to minimize the chance of hallucination. More specifically, the gradient for the least-squares component in (\ref{eq:leastsquare}) is computed for the $m$-th image estimate $\mathbf{y}_i^{m}$:
\begin{equation}
  \mathbf{\nabla}_i^{m} = \mathbf{B}_i\adj \mathbf{B}_i \mathbf{y}_i^{m} - \mathbf{B}_i\adj \mathbf{W}\mathbf{u}_i.\label{eq:gradient}
\end{equation}
Matrix $\mathbf{B}_i$ applies the forward model $\mathbf{A}_i$ for patch $i$ along with phase $e^{j2\pi(\mathbf{k}_i\cdot\mathbf{x})}$, k-space subsampling operation with matrix $\mathbf{M}_i$, and weighting $\mathbf{W}$:
\begin{equation}
  \mathbf{B}_i \mathbf{y}_i^m = \mathbf{W}\mathbf{M}_i\mathbf{A}\left(e^{j2\pi(\mathbf{k}_i\cdot\mathbf{x})} \ast \mathbf{y}_i^{m} \right).
\end{equation}
The adjoint of $\mathbf{B}_i$ is denoted as $\mathbf{B}_i\adj$. Original k-space measurements (network input) are denoted as $\mathbf{u}_i$. The gradient $\mathbf{\nabla}_i^{m}$ from (\ref{eq:gradient}) is used to update the current estimate as
\begin{equation}
  \mathbf{y}_i^{m+} = \mathbf{y}_i^{m} + t \mathbf{\nabla}_i^{m}.
\end{equation}
Different algorithms can be used to determine the step size $t$ \cite{Daubechies2004,Elad2007,Figueiredo2003,Starck2005}. For a fixed number of iterations, the optimal step size $t$ must be determined. Here, we initialize the step size $t$ to -2, and we learn a different step size for each iteration as $t^m$ to increase model flexibility.

The \textit{de-noising block} consists of a number of 2D convolutional layers to effectively de-noise $\mathbf{y}_i^{m+}$. The input image consists of 2 channels, since the real and imaginary components for complex data $\mathbf{y}_i^{m+}$ are treated as 2 separate channels. This tensor is passed through an initial convolutional layer with $3\times3$ kernels that expands the data to 128 feature maps. The data tensor is then passed through 5 layers of repeated $3\times3$ convolutional layers with 128 feature maps. A final $3\times3$ convolutional layer combines the 128 feature maps back to 2 channels. For a residual-type structure, the input to the de-noising block is added to the output. Batch normalization \cite{Ioffe2015} and Rectified Linear Unit (ReLU) layers are used after each convolutional layer except the last one. Linear activation is applied at the last layer to ensure that the sign of the data is preserved. Convolutional layers are applied using circular convolutions. MR data are acquired in the frequency domain, and the Fourier transform operator assumes that the object of interest is repeated in the image domain. The final tensor with 2 channels is then converted to complex data as the updated image $\mathbf{y}_i^{m+1}$.

The two blocks, update and de-noising, are repeated as ``iterations.'' Convolutional layer weights in the de-noising block can be kept constant for each repeated block or varied. In our experiments, weights are varied for each block. Additionally, a hard data projection is performed as a final step in the network: known measured samples are inserted into the corresponding k-space location.

\subsection{Computation}

To solve the inverse problem in \ref{eq:leastsquare}, iterative algorithms are typically used. During each iteration, inverse and forward multi-dimensional Fourier transforms are performed. Despite algorithmic advancements, the Fourier transform is still the most computationally expensive operation. For the conventional approach of reconstructing the entire 2D image at once, each Fourier transform requires $O\left(N_z N_y \log (N_yN_z)\right)$ operations for an $N_y \times N_z$ image. In our proposed approach, the reconstruction is only performed for localized patches of k-space; thus, all operations including the Fourier transform are performed with smaller image dimensions which significantly reduces computation. For example, given initial image dimensions of $N_y = 256$ and $N_z = 256$, we can perform the reconstruction as solving the inverse problem for $64 \times 64$ patches. In this case, we reduce the order of computation for the Fourier transform by over 28 fold. In practice, many other factors contribute to the reconstruction time, including reading/writing and data transfer. The proposed framework provides a powerful degree of freedom to optimize for faster reconstructions.

In the proposed design, we can further accelerate the reconstruction on two fronts. First, the reconstruction of each individual k-space patch can be performed independently. This property enables parallelization of the reconstruction process. The entire reconstruction can be performed in the time in takes to reconstruct a single patch which further highlights the savings from applying the Fourier transform on smaller image dimensions. Second, conventional iterative approaches to solve (\ref{eq:leastsquare}) require an unknown number of iterations for convergence and the need to empirically tune regularization parameters. With the proposed ISTA-based network, the number of iterations is fixed, and the network is trained to converge in the given number of steps.

\begin{figure*}[!t]
  \centering
  \includegraphics[width=\linewidth]{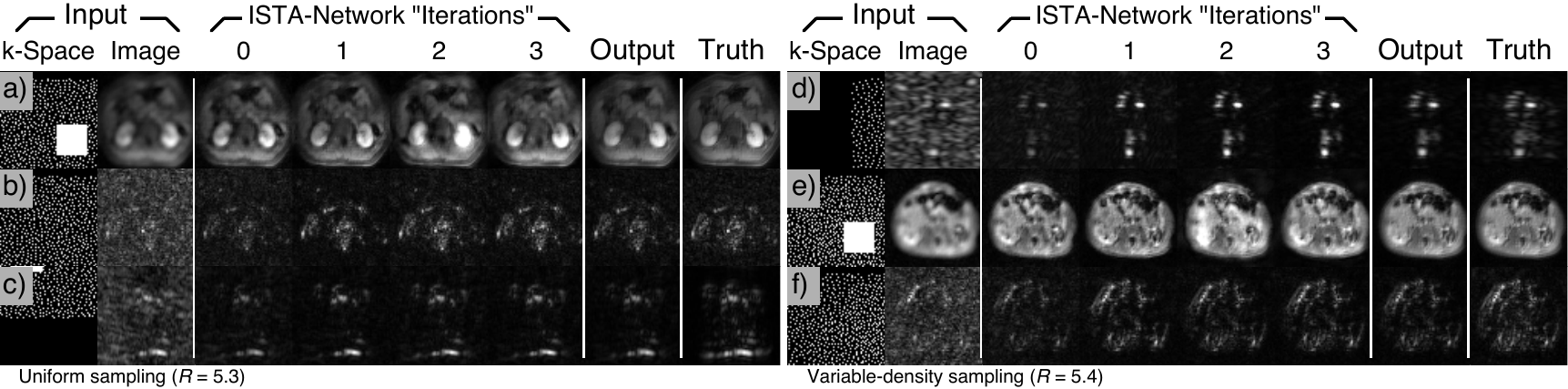}
  \caption{Representative output from randomly selected bandpass-filtered input for uniform ((a)--(c)) and variable-density ((d)--(f)) poisson-disc subsampling. Input k-space data were padded (noted by the black bands in k-space) to the same $64 \times 64$ input size, and the data was passed through a ISTA-based neural network with 4 ``iteration'' blocks. Since square frequency bands were selected, the corresponding data in the image domain had a square aspect ratio. The final output with a final hard data projection stage has comparable results to the ground truth (last two columns).}
  \label{fig:results:bandpass}
\end{figure*}

\section{Experiment Setup}\label{section:experiment}

With Institutional Board Review approval and informed consent, abdominal images were acquired using gadolinium-contrast-enhanced MRI with GE MR750 3T scanners. Both 20-channel body and 32-channel cardiac coil arrays were used. Free-breathing T1-weighted scans were collected from 301 pediatric patient volunteers using a 1--2 minute RF-spoiled gradient-recalled-echo sequence with pseudo-random Cartesian view-ordering and intrinsic motion navigation \cite{Cheng2013a,Zhang2013a}. Each scan acquired a volumetric image with a minimum dimension of $224 \times 180 \times 80$. Data were fully sampled in the $k_x$ direction (spatial frequency in $x$) and were subsampled in the $k_y$ and $k_z$ directions (spatial frequencies in $y$ and $z$). The raw imaging data were first compressed from the 20 or 32 channels to 6 virtual channels using a singular-value-decomposition-based compression algorithm \cite{Zhang2012a}.
Images were modestly subsampled with a reduction factor of 1 to 2, and images were first reconstructed using compressed-sensing-based parallel imaging. Sensitivity maps for parallel imaging were estimated using ESPIRiT \cite{Uecker2013}. Compressed sensing regularization was applied using spatial wavelets \cite{Lustig2007a}. Image artifacts from respiratory motion were suppressed by weighting measurements according to the degree of motion corruption \cite{Cheng2013a,Johnson2011}.

For training, all volumetric data were first transformed into the hybrid $(x, k_y, k_z)$-space. Each $x$-slice was considered as a separate training example. Data were divided by patient: 229 patients for training (44,006 slices), 14 patients for validation (2,688 slices), and 58 patients for testing (11,135 slices). Seventy two different sampling masks were generated using pseudo-random poisson-disc sampling \cite{Lustig2007a} with reduction factors ranging from 2 to 9 with a fully sampled calibration region of $20\times20$ in the center of the frequency space. Both uniform and variable-density sampling masks were generated. Sensitivity maps for the data acquisition model were estimated from k-space data in the calibration region using ESPIRiT \cite{Uecker2013}. As suggested in Ref.~\cite{Uecker2013}, 2 sets of ESPIRiT maps were used which resulted in the input and output of the de-noising block as a tensor with 4 channels: 2 ESPIRiT maps with complex data that were separated into 2 real and 2 imaginary channels. Since these 2 maps were highly correlated, we maintained the use of 128 feature maps in the de-noising block.

Each training example was normalized by the square root of the total energy in the center $5 \times 5$ block of k-space data. The example was then scaled by $10^5$ so that maximum pixel values in the image domain for a $64 \times 64$ patch will be on the order of 100. The Adam optimizer \cite{Kingma2014} was used with $\beta_1 = 0.9$, $\beta_2 = 0.999$, and a learning rate of 0.01 to minimize the $\ell_1$ error of the output compared to the ground truth. For each training step, a batch of random data examples were selected, and random k-space subsampling masks were applied. Afterwards, the training examples were randomly cropped to the desired k-space patch dimension.

We evaluated three main features: 1)~number of iteration blocks in the ISTA-based network, 2)~dimensions of each k-space patch, and 3)~amount of overlap between neighboring patches. First, we evaluated the impact of the number of iteration blocks in the ISTA-based network by training and applying different networks with 2, 4, 8, and 12 iteration blocks. Second, separate networks were trained for different patch dimensions: $32\times32$, $48\times48$, $64\times64$, and $80\times80$. The weights for each network were then applied to reconstruct images with varying size patches to evaluate how well the weights generalize. Additionally, we evaluated the reconstruction time as function of patch dimension. Assuming the ability to parallelize an unlimited number of patches, a single patch of varying dimensions was reconstructed 50 times, and the average inference time was reported. Third, the amount of overlap between neighboring reconstructed patches was evaluated. For simplicity, we used a constant 50\% overlap in the $k_z$ dimension and varied the amount of overlap in the $k_y$ dimension. If unspecified, experiments were performed using a patch size of $64\times64$ with a 50\% overlap, a variable-density subsampling with reduction factors of $R$ = 5.4$\pm$0.2, and ISTA-based network with 4 iterations. The final reconstructed k-space image is transformed to the image domain by applying the adjoint imaging model $\textbf{A}\adj$.

When applicable, results were compared with the subsampled input data that was reconstructed by directly applying $\textbf{A}\adj$. Also, state-of-the-art compressed-sensing reconstructions with parallel imaging and spatial Wavelet regularization were performed for comparison. Reconstructions were evaluated using peak-signal-to-noise-ratio (PSNR), root-mean-square-error normalized by the norm of the reference (NRMSE), and structural-similarity metric \cite{Wang2004d} (SSIM).

The proposed method was implemented in Python with TensorFlow\footnote{\url{https://github.com/jychengmri/bandpass-convnet}} \cite{Abadi2016}. Sensitivity map estimation with ESPIRiT, compressed sensing reconstruction, and generation of poisson-disc sampling masks were performed using the Berkeley Advanced Reconstruction Toolbox (BART)\footnote{\url{https://github.com/mrirecon/bart}} \cite{UeckerISMRM2015}.

\begin{figure*}[!t]
  \begin{center}
    \includegraphics[width=\linewidth]{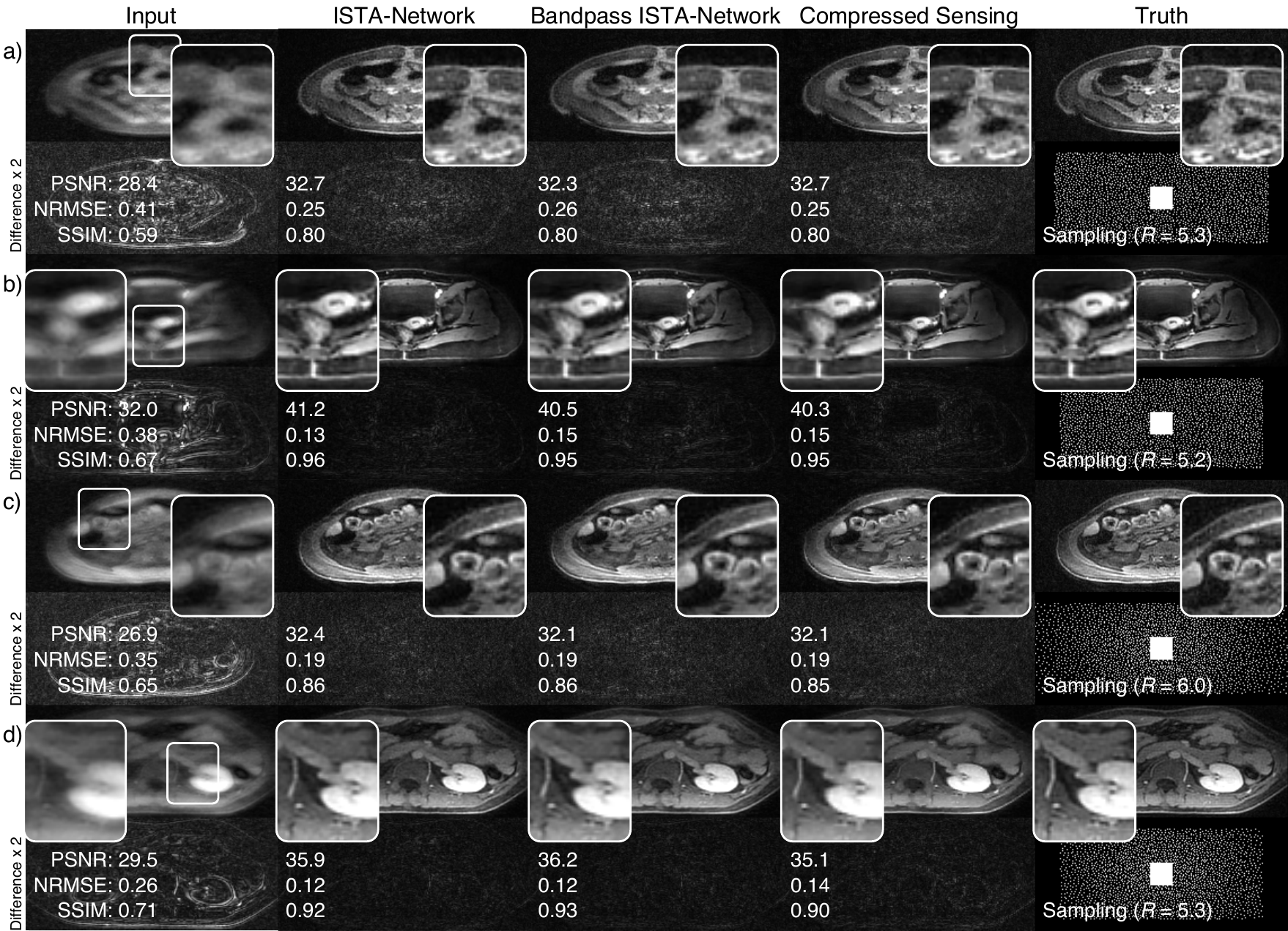}
  \end{center}
  \caption{Representative results. Test data (upper right) were subsampled with uniform ((a) and (b)) and variable-density ((c) and (d)) sampling masks (lower right) to generate the input data (left column). Images were reconstructed with a network trained on the entire image (second column) and with the proposed deep network with bandpass filtering (third column). Compressed-sensing-based parallel imaging reconstructions are also displayed (fourth column).}
  \label{fig:results:example}
\end{figure*}

\section{Results}\label{section:results}

Representative tests results for different frequency bands are shown in Fig.~\ref{fig:results:bandpass}. The final results are comparable with compressed sensing in Fig.~\ref{fig:results:example}.

\begin{table}[!t]
  \caption{Reconstruction Performance}
  \label{table:results:iteration}
  \small
  \centering
  \begin{tabular}{|l|c|c|c|}
    \hline
    Method & PSNR & NRMSE & SSIM\\
    \hline\hline
    \multicolumn{4}{|l|}{Uniform subsampling ($R$ = 5.3$\pm$0.1)} \\
    \hline
    Input & 29.3$\pm$2.3 & 0.35$\pm$0.09 & 0.67$\pm$0.08 \\
    BP-Net x2 & 33.0$\pm$2.7 & 0.23$\pm$0.07 & 0.83$\pm$0.07 \\
    BP-Net x4 & 34.6$\pm$3.1 & 0.19$\pm$0.07 & 0.87$\pm$0.07 \\
    BP-Net x8 & 35.0$\pm$3.3 & 0.18$\pm$0.06 & 0.87$\pm$0.06 \\
    BP-Net x12 & 35.3$\pm$3.4 & 0.18$\pm$0.06 & \textbf{0.88$\pm$0.06}\\
    Net x4 & 35.3$\pm$3.4 & 0.18$\pm$0.06 & 0.87$\pm$0.06 \\
    Compressed sensing & \textbf{35.6$\pm$3.6} & \textbf{0.17$\pm$0.06} & 0.87$\pm$0.06 \\
    \hline\hline
    \multicolumn{4}{|l|}{Variable-density ($R$ = 5.4$\pm$0.2)} \\
    \hline
    Input & 29.5$\pm$2.3 & 0.34$\pm$0.09 & 0.67$\pm$0.08 \\
    BP-Net x2 & 33.8$\pm$2.9 & 0.21$\pm$0.06 & 0.85$\pm$0.07 \\
    BP-Net x4 & 35.5$\pm$3.3 & 0.18$\pm$0.06 & 0.88$\pm$0.06 \\
    BP-Net x8 & 35.8$\pm$3.5 & 0.17$\pm$0.06 & \textbf{0.89$\pm$0.06} \\
    BP-Net x12 & \textbf{36.1$\pm$3.6} & \textbf{0.16$\pm$0.06} & \textbf{0.89$\pm$0.06} \\
    Net x4 & 36.0$\pm$3.6 & 0.17$\pm$0.06 & 0.88$\pm$0.06 \\
    Compressed sensing & 36.0$\pm$3.7 & 0.17$\pm$0.06 & 0.88$\pm$0.06 \\
    \hline
  \end{tabular}
\end{table}

The impact of the number of iteration block on reconstruction performance is summarized in  Table~\ref{table:results:iteration}. When more iteration blocks were used in the proposed bandpass network, the reconstruction performance improved with higher PSNR, lower NRMSE, and higher SSIM. The most gains were seen going from 2 iteration blocks with SSIM values of 0.83 and 0.85 to 4 iteration blocks with SSIM values of 0.87 and 0.88 for both uniform and variable-density sampling, respectively. With 12 iteration blocks, the bandpass network performed similarly to compressed sensing. To evaluate other components of the network, 4 iterations were used to balance between performance and depth.

\begin{figure}
  \centering
  \includegraphics[width=1.0\linewidth]{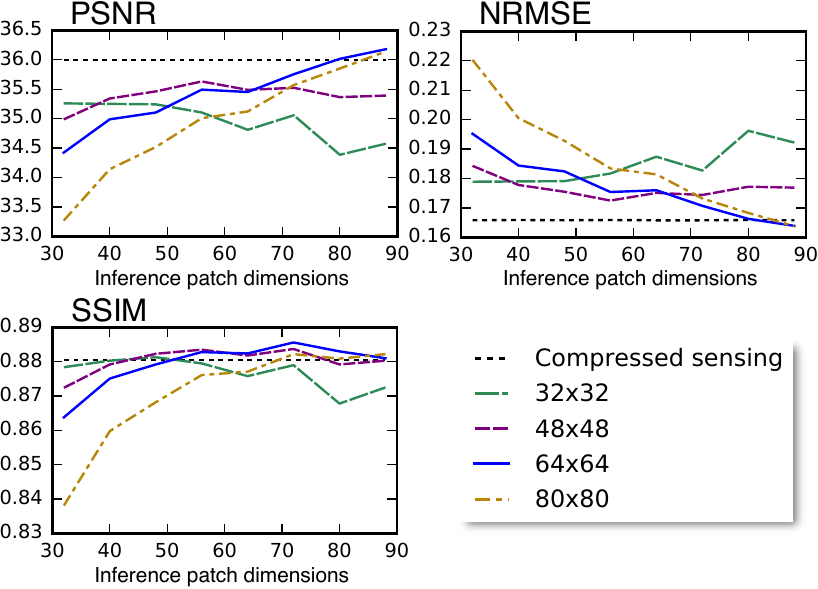}
  \caption{Performance as a function of patch dimension. The bandpass network was separately trained for specific patch dimensions: $32\times32$ (long-dash green), $48\times48$ (short-dash purple), $64\times64$ (blue), and $80\times80$ (dash-dot orange). During testing, weights trained were applied for varying patch dimensions. Compressed sensing (dot black) does not use the patch dimension and is plotted for reference.}
  \label{fig:results:bpshape}
\end{figure}

The impact of patch dimensions is shown in Fig.~\ref{fig:results:bpshape}. Reconstruction performance improved for larger patch dimensions during inference. By training the bandpass network specifically for smaller patch dimensions ($32\times32$), the reconstruction performance was best for smaller patch dimensions during inference. However, maximum PSNR and SSIM were lower and minimum NRMSE was higher for this bandpass network compared to the bandpass network trained and applied with larger patch dimensions. For all cases, the trained network can be applied to a small range of different patch dimensions. The $(64\times64)$-trained bandpass network had improved performance in terms of SSIM for inference on $70\times70$ patches, but performance began to degrade for inference on patches larger than $80\times80$. The $(48\times48)$-trained network had similar performance to the $(64\times64)$-trained network but with maximum SSIM shifted towards smaller patch dimensions.

\begin{figure}
  \centering
  \includegraphics[width=0.6\linewidth]{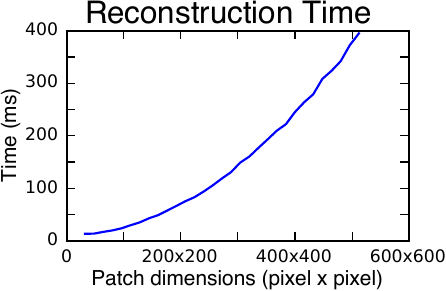}
  \caption{Reconstruction (inference) time for a single patch as a function of patch dimension on a NVIDIA Titan X card. A single patch of the specified dimensions was reconstructed 50 times with a ISTA-based neural network built with 4 iterations, and the average inference time is plotted. The total reconstruction time increased quadratically with respect to patch dimension.}
  \label{fig:results:time}
\end{figure}

The impact of patch dimension on reconstruction time is summaried in Fig.~\ref{fig:results:time}. In this plot, average inference time to reconstruct a single patch for 50 runs is plotted with respect to the patch size. The main advantage of the approach is its the ability to parallelize the reconstruction. If the entire image was considered as a single patch, the average time to reconstruct a single $512\times512$ image was 395~ms. A single $64\times64$ patch was reconstructed with the trained CNN in 17~ms --- a 23-fold speed up in reconstruction time. With enough computation resources, this gain can be realized by reconstructing the $512\times512$ image as $64\times64$ k-space patches.

\begin{figure}
  \begin{center}
    \includegraphics[width=\linewidth]{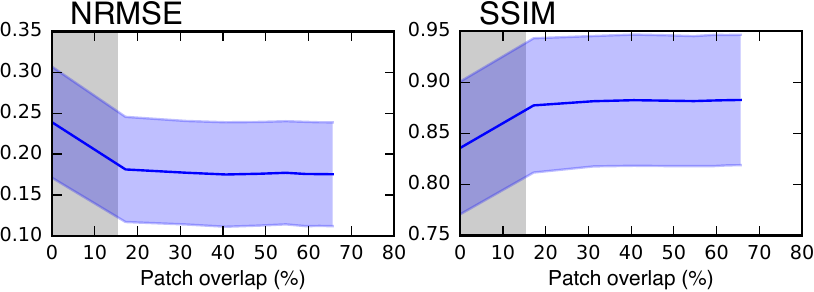}
  \end{center}
  \caption{Reconstruction performance as a function of percent of overlap between neighboring patches in the $y$ dimension. For the $64\times64$ block, the window function had a passband of $44\times44$. To cover the stopband of the window function, a minimum of 15.6\% overlap was needed (highlighted in gray). After 20\%, performance was relatively immune to amount of overlap. Same trends were observed for PSNR (not shown).}
  \label{fig:results:overlap}
\end{figure}

The impact of overlap between neighboring patches is summarized in Fig.~\ref{fig:results:overlap}. Loss of performance was noted if the amount of overlap is less than the stopband of the window function. In this case, either part of the k-space was not reconstructed, or errors near the stopband of the window function were accentuated. Above an overlap threshold of around 15\%, NRMSE and SSIM were relatively independent to changes in amount of patch overlap. Fewer patches can be reconstructed by minimizing the amount of overlap between neighboring patches. Conversely, more patch overlap yielded negligible gains. Therefore, for computational efficiency and without any loss in accuracy, the patch size should be set to the minimal size needed to account for the window stopband.

\begin{figure}[!t]
  \begin{center}
    \includegraphics[width=\linewidth]{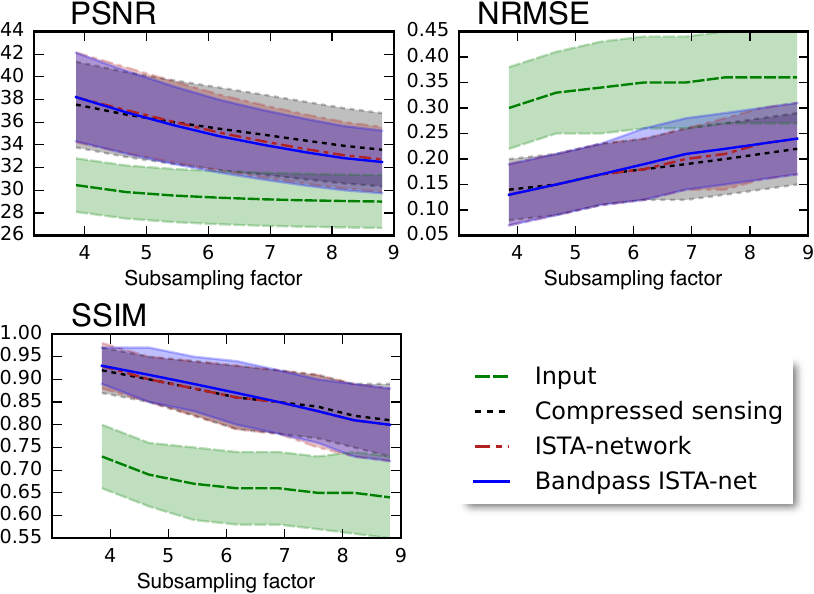}
  \end{center}
  \caption{Reconstruction performance as a function of subsampling factor for variable-density sampling. Standard deviations for each value are illustrated with transparent color shading. The proposed bandpass network (blue), full network (dash-dot red), and compressed sensing (dashed black) have near identical performance in terms of PSNR, NRMSE, and SSIM.}
  \label{fig:results:acceleration}
\end{figure}

The effect of subsampling factor ($R$) on reconstruction performance is shown in Fig.~\ref{fig:results:acceleration}. The bandpass network with $64\times64$ patches, $50\%$ overlap, and 4 iteration blocks was trained with both uniform and variable-density subsampling ($R$ = 2--9). Overall, higher subsampling factors resulted in lower PSNR and SSIM and higher NRMSE. The proposed method performed comparably to compressed sensing and with a network trained specifically on the full image. Slight discrepancies may be the result of an imbalance of subsampling factors and patterns during training. Similar trends were observed for uniform subsampling (not shown) with minor loss in performance for the same subsampling factors as seen in Table~\ref{table:results:iteration}.

\section{Discussion}\label{section:discussion}
We introduced the use of bandpass filtering to enable parallelization of the image reconstruction while maintaining the use of the data acquisition model. We developed and demonstrated this approach in a deep-learning framework. The data-driven strategy with deep learning eliminates the need to engineer priors by hand for each frequency band and enables generalization of this approach to different applications.

An unrolled network based on ISTA was used as the core network. The setup can be easily adapted for more sophisticated network architecture. Also, the training can include loss functions that correlate better with diagnostic image quality such as with a generative adversarial network \cite{Goodfellow2014,Lohit2017,Mardani,Quan2017}. For simplicity, we chose to implement the network for complex numbers as 2 separate channels, and we were able to demonstrate high image quality. The network can be further improved by considering complex data in each operation of the neural network \cite{Trabelsi2017,Virtue2017}.

\begin{figure}
  \begin{center}
    \includegraphics[width=1.0\linewidth]{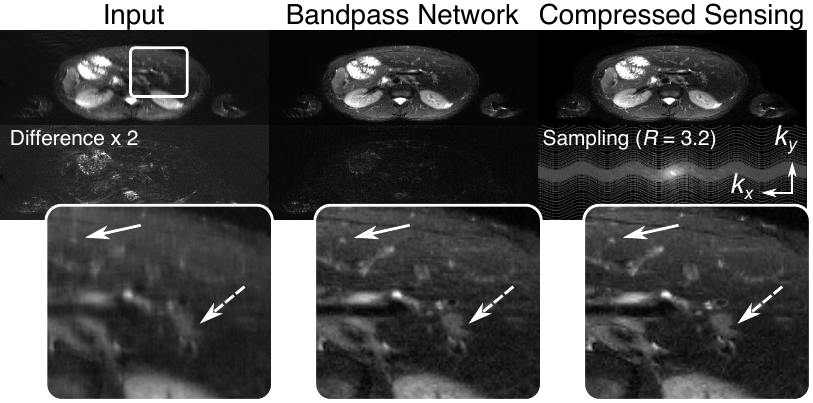}
  \end{center}
  \caption{T2-weighted 2D abdominal scan with wave sampling. The proposed technique was adapted to support wave-encoding (subsampling in middle right). Input (top left) and bandpass ConvNet output (top middle) are displayed along with the difference with the compressed sensing reconstruction (right). Enlarged images (bottom row) highlight recovery of fine details (arrows).}
  \label{fig:results:wave}
\end{figure}

An advantage of the network structure is the ability to include more sophisticated imaging models. For example, non-Cartesian sampling trajectories offer the ability to reduce MRI scan durations even before subsampling the acquisition. To demonstrate this flexibility, we applied the bandpass network to hybrid Cartesian MRI. More specifically, we applied our approach to wave-encoded imaging \cite{Bilgic2014,Chen2017,Moriguchi2006}. In this case, sinusoids were used for the k-space sampling trajectory. We adapted the imaging model $\textbf{A}$ to include an operator that grids the non-Cartesian sampling onto a Cartesian grid \cite{Bilgic2014,Chen2017}. Multi-slice 2D T2-weighted single-shot fast-spin-echo abdominal scans were acquired from 137 patient volunteers on a 3T scanner with a subsampling factor of 3.2. Data were divided as 104 patients (5005 slices) for training, 8 patients (383 slices) for validation, and 25 patients (1231 slices) for testing. Due to T2 signal decay and patient motion, fully sampled datasets cannot be obtained; thus, the ground truth was obtained through a compressed-sensing reconstruction for wave encoding \cite{Chen2017}. Though the ground truth was biased towards the compressed-sensing reconstruction, we still demonstrated the ability of the bandpass network to reconstruct more general imaging models. In Fig.~\ref{fig:results:wave}, the bandpass network method was able to recover image sharpness and yielded comparable results to compressed sensing.

The output of our proposed network had the same number of input complex data channels. This property enabled the ability to replace the estimated samples with original measurements. In doing so, the final reconstructed image will not deviate from the measured samples. Furthermore, if there are concerns with diagnostic accuracy of CNN results, the estimated data can be weighted down. Alternatively, the output can be easily used as initialization for conventional approaches.


To reduce the input dimensions and to enable parallelization, each localized patch of data was reconstructed independently. One possible limitation with the proposed architecture was that not all image properties was explicitly exploited. Different k-space patches could be highly correlated and could assist in the reconstruction of other patches. For instance, if the final image is assumed to be real valued, the frequency-space image should have hermitian symmetry. We hypothesize that the deep network was able to model and infer some of these characteristics. Complementary information may already be implicitly embedded in the input data: signal amplitude and image structure type may indicate patch location. If needed, the infrastructure can be easily extended to include additive information. Complementary information can be included in the input to the de-noising block.

Another imaging property that was not leveraged in the current work was the specific correlation properties for different frequency bands \cite{Gong2014,Li2017}. Here, we simplified the setup by training a single ConvNet that can be applied to reconstruct any frequency patch, and we rely on the flexibility of the nonlinear model to adapt to different patches. When we reduced the training patch dimensions, the number of different frequency bands increased. As a result, the required model size and the training duration also increased due to the need to model a wider variation of features. Future work includes investigating the gains of applying different models for patches from different frequency regions.

\section{Conclusion}
A bandpass deep neural network architecture was developed and demonstrated here to solve the inverse problem of estimating missing measurements of subsampled MRI datasets. The main advantages of the bandpass network were leveraged when the division of data into localized patches was performed in the measurement domain. The highly scalable and flexible architecture can be adapted for other applications in MRI, such as detection and correction of corrupt measurements on a patch-by-patch basis. Additionally, this approach can be adapted for other applications, such as super-resolution or image de-noising. Working in the hybrid frequency-spatial-space offers unique image-processing properties that can be further investigated.

\section*{Acknowledgements}
We are grateful for the support from NIH R01-EB009690, NIH R01-EB019241, NIH R01-EB026136, and GE Healthcare.

\ifCLASSOPTIONcaptionsoff
  \newpage
\fi



\bibliographystyle{IEEEtran}
\bibliography{library}

\end{document}